\documentclass[runningheads]{llncs}
\usepackage{graphicx}
\usepackage{amsmath} 
\usepackage{amssymb}
\usepackage{colortbl}
\usepackage{multirow}
\usepackage{color}
\usepackage{tabularray}
\usepackage{adjustbox}
\usepackage{hhline}
\usepackage{colortbl}
\usepackage{color}
\usepackage[misc]{ifsym}

\usepackage{hyperref}
\hypersetup{
    colorlinks=true,
    linkcolor=blue,
    filecolor=magenta,      
    urlcolor=blue,
    pdftitle={Overleaf Example},
    pdfpagemode=FullScreen,
    }

\newcommand\blfootnote[1]{%
  \begingroup
  \renewcommand\thefootnote{}\footnote{#1}%
  \addtocounter{footnote}{-1}%
  \endgroup
}

\usepackage[dvipsnames]{xcolor}
\begin{document}
%
%\title{Deform3DGS: Fast Reconstruction of Endoscopic Scenes via FlexiDeform}
\title{Deform3DGS: Flexible Deformation for \\ Fast Surgical Scene Reconstruction \\ with Gaussian Splatting }
\titlerunning{Flexible Deformation for Fast Surgical Scene Reconstruction}
% \author{Paper ID: 3887}
\institute{}
\author{Shuojue Yang\inst{1} \and
Qian Li\inst{1} \and
Daiyun Shen\inst{2}\ \and
Bingchen Gong\inst{3} \and
Qi Dou\inst{3} \and
Yueming Jin\inst{1}\textsuperscript{(\Letter)}
}
\authorrunning{S. Yang et al.}
\institute{National University of Singapore, Singapore, Singapore \and
Tsinghua University, Bejing, China \and
 The Chinese University of Hong Kong, Hong Kong, China \\
\email{ymjin@nus.edu.sg}\\
}

\maketitle              % typeset the header of the contribution
\begin{abstract}
Tissue deformation poses a key challenge for accurate surgical scene reconstruction. Despite yielding high reconstruction quality, existing methods suffer from slow rendering speeds and long training times, limiting their intraoperative applicability. 
\blfootnote{\textsuperscript{} D. Shen did this work during his internship at National University of Singapore.}
Motivated by recent progress in 3D Gaussian Splatting, an emerging technology in real-time 3D rendering, this work presents a novel fast reconstruction framework, termed \emph{\textbf{Deform3DGS}}, for deformable tissues during endoscopic surgery. Specifically, we introduce 3D GS into surgical scenes by integrating a point cloud initialization to improve reconstruction. Furthermore, we propose a novel flexible deformation modeling scheme (FDM) to learn tissue deformation dynamics at the level of individual Gaussians. Our FDM can model the surface deformation with efficient representations, allowing for real-time rendering performance. More importantly, FDM significantly accelerates surgical scene reconstruction, demonstrating considerable clinical values, particularly in intraoperative settings where time efficiency is crucial. Experiments on DaVinci robotic surgery videos indicate the efficacy of our approach, showcasing superior reconstruction fidelity PSNR: (37.90) and rendering speed (338.8 FPS) while substantially reducing training time to only 1 minute/scene. Our code is available at \url{https://github.com/jinlab-imvr/Deform3DGS}.

\end{abstract}

\section{Introduction}
% why 3D Reconstruction in Laparoscopic Surgery
% Endoscopic surgery has now dominated many abdominal surgical procedures with minimal risk of infections and improved clinical outcomes~\cite{}. 
%Recently, there has been a growing research interest in the 
Three-dimensional (3D) reconstruction of surgical scenes has great potential to facilitate many downstream applications including intraoperative navigation~\cite{pelanis2021evaluation}, and visualization enhancement~\cite{rodby2014advances,liu2020reconstructing}. Besides, a high-quality and dynamic 3D scene model has demonstrated potential benefits to surgical training through shortening learning curves~\cite{boedecker2021using} and remote surgical proctoring~\cite{wu2023amassing} by allowing immersive observation of surgical scenes.   
% Driven by the large clinical significance, we study the efficient reconstruction of deformable tissues in this work given stereo endoscopic videos.

% line of NeRF-based works: why, limitations and related works
%ray marching, a numeric integration technique to render scenes
Conventional reconstruction methods suffer from a redundant workflow including depth estimation, surface reconstruction, and texture mapping~\cite{schonberger2016structure}. To enhance the compactness and efficiency, many studies introduce Neural Radiance Field (NeRF)~\cite{mildenhall2021nerf}, an implicit 3D representation using tiny Multi-Layer Perceptrons (MLPs), to model geometric details and appearance from the captured video. This implicit representation can directly render the photo-realistic novel views, simplifying the bulky conventional workflow. Works~\cite{wang2022neural,zha2023endosurf} successfully adapt NeRF into endoscopic scene reconstruction, leading to promising performance in rendering quality and geometry fidelity. However, this line of work suffers from a long training time (hours) and low rendering speed, which significantly impedes their intraoperative applicability. LerPlane~\cite{yang2023neural} encodes the spatial and temporal information using decomposed 4D feature planes to accelerate the reconstruction. This approach inherently models the deformation and directly outputs rendering parameters for deformed tissues, which heavily rely on complex computations performed on feature planes and MLPs, leading to a compromised acceleration.
%[width=0.8\textwidth]
% \begin{figure}[t]
%     \centering
%     \includegraphics[height=5cm]{deformation_fields.png}
%     \caption{Comparison of deformation models used by LerPlane, 4DGS, and our framework. $\boldsymbol x$ and $t$ denote 3D spatial coordinate and time, respectively. Note that our deformation model obviates the need for MLPs, leading to highly efficient deformation modeling and inference. The deformation is illustrated by displacement $\triangle \boldsymbol{x}$.}
%     \label{deformation_fields}
% \end{figure}

With the current progress in computer graphics, Gaussian Splatting (GS)~\cite{kerbl20233d} emerged as a ground-breaking 3D representation. Driven by its superior performance, many efforts~\cite{yang2023deformable3dgs,li2023spacetimegaussians,kratimenos2023dynmf} have been made to adapt this technique to dynamic scene reconstruction through a deformation model to represent motions. 4DGS~\cite{wu20234d} is one of the pilot studies presenting one efficient solution where decomposed feature planes similar to ~\cite{cao2023hexplane} are used to model time-dependent deformations. Despite the compact feature encoding, 4DGS retains time-consuming feature plane interpolation and decoding with MLPs, which limits the acceleration and intraoperative applicability. Also, 4DGS inadequately utilizes geometric priors, leading to a highly sparse point initialization and a longer time for geometry reconstruction. Therefore, although exhibiting large values for fast reconstruction on dynamic scenes, integrating GS into the surgical scene reconstruction framework is still challenging, which inspires our work toward effectively adapting GS to intraoperative scenes.
%Benefiting from stereo endoscopes used during robot-assisted endoscopic surgery,
% Given the estimated pixel-wise stereo depth~\cite{}, our MAPF can densely and reliably initialize point cloud.
% Furthermore, to address local sparsity resulting from surgical instrument occlusions or invalid depth estimations, MAPF selectively integrates 3D points extracted from other frames into regions exhibiting significant motion and sparsity, which enhances the point density in areas prone to occlusions or unreliable depth measurements.where 3D tissue points are selectively integrated into regions exhibiting significant motion and sparsity. 

In this paper, we develop a highly efficient framework for deformable surgical scene reconstruction, named Deform3DGS, by introducing the GS technique into surgical scenario with a motion-aware point fusion (MAPF) to initialize the Gaussian point cloud densely. Besides, we propose a novel flexible deformation modeling scheme (FDM) for efficiently representing tissue deformations. FDM models the tissue deformations via an efficient linear combination regression, where learnable basis functions are leveraged to improve both the representation capability and efficiency. Finally, we evaluate our method on EndoNeRF~\cite{wang2022neural} and StereoMIS~\cite{hayoz2023pose} datasets collected from Da Vinci robotic surgery videos. Experiments indicate the efficacy of our approach, demonstrating superior reconstruction quality (PSNR: 37.90) and rendering speed (338.8 FPS) while substantially reducing training time to around only 1 minute per scene.
% Motivated by these issues, in this work, we propose a novel framework for the fast reconstruction of deformable endoscopic scenes. Given the stereo depth, we adopt the point cloud recovered from the first frame as the initialization. However, endoscopic images inevitably contain occlusions and blurs where the recovered initial point cloud unevenly distributes, leading to a longer optimization time. Therefore, we propose a Motion-aware Point Fusion (MAPF) strategy to fuse the estimated point cloud of each frame into the initialized distribution. To densify the locally sparse distribution, 3D points at various timestamps leading to intensive motions are selected and projected to the sparse regions in canonical space. Besides, to accelerate the deformation optimization while maintaining the quality, we propose a novel Gaussian-driven deformation Model (GDDM) where Gaussian radial basis functions (RBFs) with learnable parameters are linearly combined to compute the position, rotation, and scale deformations for each Gaussian point. Of note, our model obviates the need for training MLPs and encoding/decoding procedures, which drastically improves the rendering and training efficiency. We evaluate our method on EndoNeRF and StereoMIS datasets collected from Da Vinci robotic surgery videos. Experiments indicate the efficacy of our approach, showcasing superior reconstruction fidelity (PSNR: 37.76) and rendering speed (319 FPS) while substantially reducing training time to only 1 minute/scene.\textcolor{red}
\section{Method}
\begin{figure}[t]
    \centering
    \includegraphics[width=0.93\textwidth]{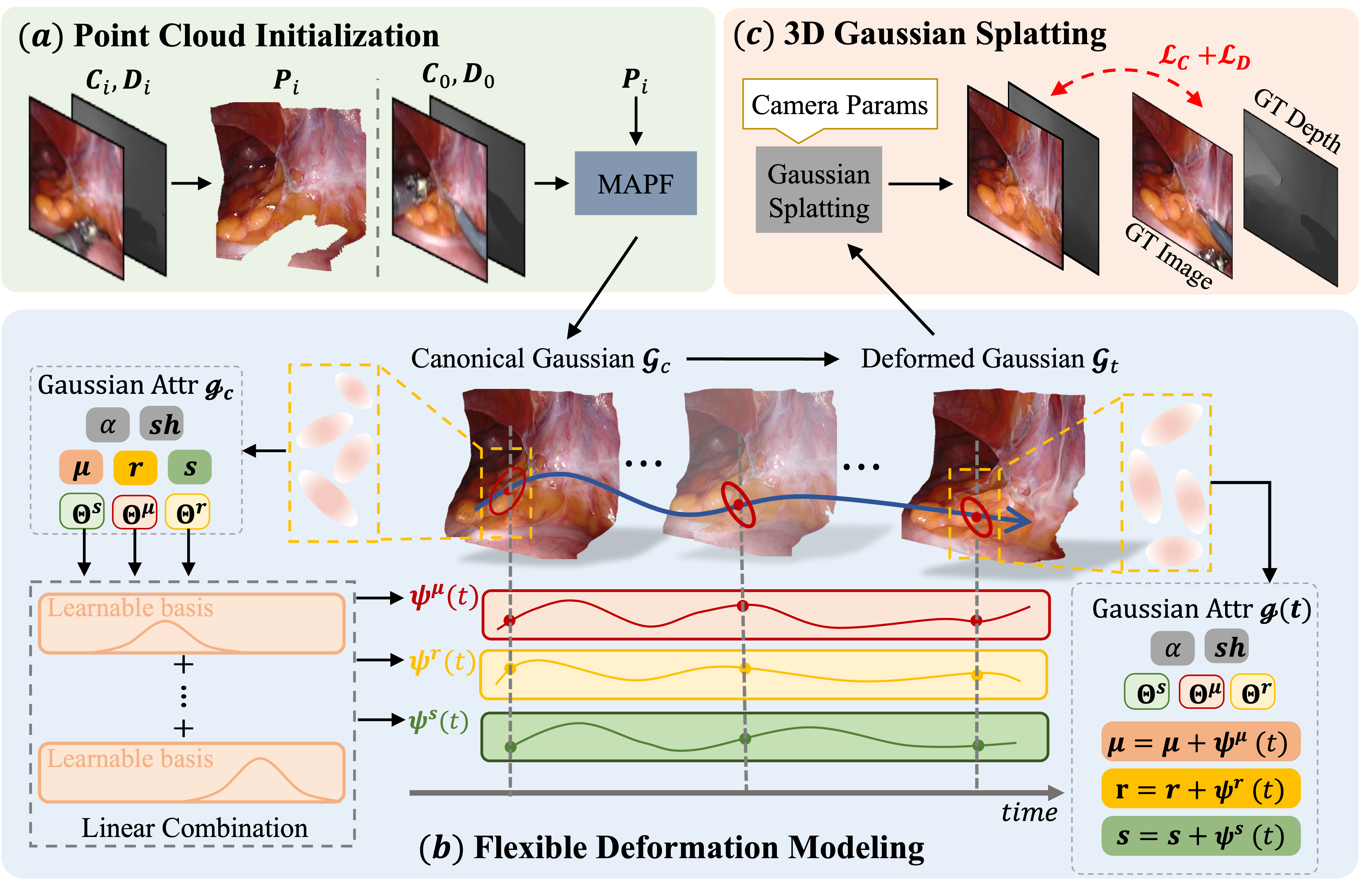}
    \caption{Illustration of our fast deformable tissue reconstruction framework, Deform3DGS, composed of (a) Point cloud initialization, (b) Flexible Deformation Modeling, and (c) 3D Gaussian Splatting.}
    \label{framework}
\end{figure}

\noindent\textbf{Pipeline.} As shown in Fig.~\ref{framework}, during the training phase, our framework initializes the endoscopic scene with a Gaussian point cloud $\mathcal{G}$ using the MAPF scheme ({Sec.~\ref{MAPF}}). The following is the FDM that models deformed tissues by learning the time-dependent changes for each Gaussian point $\mathfrak{g}$ ({Sec.~\ref{sec_FDM}}). Next, the rendering ({Sec.~\ref{sec_rendering}}) is performed to obtain the colored image and depth of the deformed Gaussian point cloud given the camera viewpoint. Finally, the rendered image and depth are supervised by corresponding ground-truth (GT) image and stereo depth to optimize the framework (Sec.2.5). During rendering (testing), given a query time and camera viewpoint, FDM deforms the learned Gaussian point cloud $\mathcal{G}$ and the view of the deformed Gaussians is rendered.

%Here we select the left view as $\boldsymbol{C}$ and stereo matching algorithms~\cite{} are performed to estimate $\boldsymbol{D}$. $\boldsymbol{M}$ indicates pixels belonging to tissues with valid depth and filters out occluded pixels. $\boldsymbol{T}$ denotes the current camera pose in the world coordinate frame. In addition, for each video, the intrinsic parameter of cameras has been calibrated and denoted by $\boldsymbol{K}$.
\noindent\textbf{Problem Setting.} Our task is to train a deformable tissue reconstruction model $\boldsymbol\Psi$ from an endoscopic surgery video. Given the camera intrinsic matrix $\boldsymbol{K}$ and extrinsic matrix $\boldsymbol{T}_i$ recording the camera viewpoint information at the $i$-th frame, a desired reconstruction model $\boldsymbol\Psi^\ast$ is supposed to render the $i$-th view $\boldsymbol{I}_i$ at timestamp $t_i$ as $\boldsymbol{I}_i = \boldsymbol\Psi^\ast(\boldsymbol{T}_i, t_i; \boldsymbol{K})$.

\subsection{Preliminaries of Gaussian Splatting}
\label{sec_rendering}
3D Gaussian Splatting~\cite{kerbl20233d} is a static 3D scene representation that models scenes with the form of a 3D Gaussian point cloud in the world coordinate frame. Each Gaussian point contains learnable attributes: position $\boldsymbol{\mu}\in\mathbb{R}^{3}$, rotation $\boldsymbol{r}\in\mathbb{R}^{4}$, scale $\boldsymbol{s}\in\mathbb{R}^{3}$, opacity $\boldsymbol{\alpha}$ and spherical harmonic ($sh$) coefficients. Given an arbitrary 3D coordinate $\boldsymbol x \in\mathbb{R}^{3}$ in world frame, the spatial impact of a 3D Gaussian point on $\boldsymbol x$ is defined by a Gaussian distribution as following:
\begin{equation}
    f\mathit(\boldsymbol x\mathit;\mathit\;\boldsymbol\mu\mathit,\boldsymbol{\mathit\;}\boldsymbol\Sigma\boldsymbol{\mathit\;}\mathit)\mathit=\exp\left({\boldsymbol{\mathit-}\frac{\boldsymbol1}{\boldsymbol2}{\boldsymbol{\mathit(}\boldsymbol x\boldsymbol{\mathit\;}\boldsymbol{\mathit-}\boldsymbol{\mathit\;}\boldsymbol\mu\boldsymbol{\mathit)}}^{\boldsymbol T}\boldsymbol\Sigma^{\boldsymbol{\mathit-}\boldsymbol1}\boldsymbol{\mathit(}\boldsymbol x\boldsymbol{\mathit\;}\boldsymbol{\mathit-}\boldsymbol{\mathit\;}\boldsymbol\mu\boldsymbol{\mathit)}}\right)
\end{equation}
\begin{equation}
   \boldsymbol\Sigma=\boldsymbol R\boldsymbol S\boldsymbol S^{\mathbf T}\boldsymbol R^{\mathbf T},
\end{equation}
where $\boldsymbol  R$ is the rotation matrix calculated from $ \boldsymbol r$, and $\boldsymbol  S$ is the diagonal matrix of $ \boldsymbol s$. Next, the 3D Gaussian point is projected to the 2D image plane for rendering. The projected 2D Gaussian with the position $\boldsymbol\mu^{\mathbf2\mathbf D}$ and covariance matrix 
 $\boldsymbol\Sigma^{\mathbf2\mathbf D}$ can be analytically computed in pixel coordinate frame given the camera intrinsic and extrinsic parameters ($ \boldsymbol K$ and $\boldsymbol T$). Finally, an $\alpha$-blending~\cite{kerbl20233d} is performed to render the colored images $ \boldsymbol{\widehat C}$ and corresponding depth $ \boldsymbol{\widehat D}$. 
%  \begin{equation}
%      C(p)=\sum_{i=1}^nc_i\alpha_if(p\vert\mu^{2D},\;\Sigma^{2D})\prod_{j=1}^{i-1}(1-\alpha_jf(p\vert\mu^{2D},\;\Sigma^{2D}))
%  \end{equation}
% % and 2D covariance matrix $\boldsymbol\Sigma^{\mathbf2\mathbf D}$ is calculated by $\boldsymbol\mu^{\mathbf2\mathbf D}=\boldsymbol P\boldsymbol W\boldsymbol\mu$ and  $\boldsymbol\Sigma^{\mathbf2\mathbf D}=\boldsymbol J\boldsymbol W\boldsymbol\Sigma\boldsymbol W^{\mathbf T}\boldsymbol J^{\mathbf T}$, respectively, 
% % \begin{align}
% %     \boldsymbol\mu^{\mathbf2\mathbf D}&=\boldsymbol P\boldsymbol W\boldsymbol\mu \\
% %     \boldsymbol\Sigma^{\mathbf2\mathbf D}&=\boldsymbol J\boldsymbol W\boldsymbol\Sigma\boldsymbol W^{\mathbf T}\boldsymbol J^{\mathbf T},
% % \end{align}
% % where $\boldsymbol P$, $\boldsymbol W$, and $\boldsymbol J$ denote the projective transformation, viewing transformation, and the Jacobian approximation of transformation $\boldsymbol P$, respectively. 

% where $\boldsymbol C$ and $\boldsymbol D$ denotes rendered image and depth, respectively. $N$ is the set of sorted Gaussians overlapping with the given pixel,
\subsection{Flexible Deformation Modeling}
\label{sec_FDM}
MLP-based implicit deformation fields represented by Hexplane~\cite{cao2023hexplane} incur significant computational overhead to training and cannot meet the stringent real-time processing demands of surgical video analytics. To mitigate it, \cite{lin2023gaussian} proposed a computationally efficient explicit representation of the deformation field, in which Fourier and polynomial basis functions ${b}(t)$ parameterized by time are employed to learn the per-Gaussian motion curve. Each Gaussian is endowed with a set of learnable weights $\boldsymbol\omega$ that linearly combine the basis functions to generate the motion curve as $\textstyle\psi(t;\boldsymbol\omega)=\sum_{j=1}^{B}\omega_j {b}_j(t)$. Despite the adaptability of each Gaussian to deformation through weight adjustments, the temporal deformation is confined to canonical motion, leading to inconsistent deformation representations for different queried times. The model may be compelled to forgo certain specific movements to ensure a coherent overall trajectory, which is detrimental in scenes with intricate and nuanced deformations, such as instrument-tissue interactions. Furthermore, the deformation learned at a particular timestamp globally influences the entire trajectory, resulting in a sub-optimal local deformation representation at different queried times. 

To overcome these limitations, we introduce a novel flexible deformation modeling scheme (FDM) which offers flexibility and adaptability to basis functions using learnable parameters. We adopt the Gaussian function with learnable center $\theta$ and variance $\sigma$.
\begin{align} 
    \tilde{b}(t;\theta, \sigma)&=\exp{\left(-\frac{1}{2\sigma^2}(t-\theta)^2\right)},
    \label{gaussian_kernel}
\end{align}
For each Gaussian $\mathfrak{g}$ in the point cloud,  position $\boldsymbol\mu$ and rotation $\boldsymbol r$ are naturally related to tissue motions, and scale $\boldsymbol s$ keeps varying since tissues are prone to elastic deformations during instrument intervention. Thus, an additional set of learnable parameters $\boldsymbol\Theta^\mu$,$\boldsymbol\Theta^r$,$\boldsymbol\Theta^s$ are introduced for each Gaussian to describe temporal deformations in their position, rotation, and scale respectively, see Fig.~\ref{framework}. Taking the positional change in x direction as an example, deformation curve can be expressed by a set of parameters $\boldsymbol\Theta^{\mu,x}=\{\boldsymbol\omega^{\mu,x},\boldsymbol\theta^{\mu,x},\boldsymbol\sigma^{\mu,x}\}$ as:
\begin{align}
    \textstyle\psi^{\mu,x}(t; \boldsymbol\Theta^{\mu,x})&=\sum_{j=1}^{B}\omega^{\mu,x}_j \tilde{b} (t;\theta_j^{\mu,x},\sigma_j^{\mu,x}),
    \label{deformation}
\end{align}

Due to the locally valuable nature of Gaussian functions, our model ensures that deformations at adjacent moments remain continuous, while those across large time intervals are almost decoupled. Besides, integrating with learnable parameters, this approach yields a deformation model that is not only computationally efficient but also capable of capturing intricate deformation dynamics.

\subsection{Point Cloud Initialization}
\label{MAPF}
{To further boost the reconstruction performance and stabilize the training, we introduce a Gaussian point cloud initialization before the deformation modeling. Specifically}, we first employ the camera model and intrinsic matrix to extract the 3D tissue point clouds for each frame as:
% \begin{equation}
% {\boldsymbol P}_{ i}=\boldsymbol K^{\boldsymbol-\mathbf1}{\boldsymbol D}_{ i}{\left[{\boldsymbol{I}_i\odot \boldsymbol{M}_i\;,\;\boldsymbol{1}}\right]}^{ T},
% \end{equation}
\begin{equation}
{\boldsymbol P}_{ i}=\boldsymbol K^{\boldsymbol-\mathbf1}{\boldsymbol D}_{ i}{({\boldsymbol{I}_i\odot \boldsymbol{M}_i})},
\end{equation}
where ${\boldsymbol P}_{i}$ and ${\boldsymbol I}_{i}$ respectively denote 3D point cloud and 2D pixel coordinates from the $i$-th frame, $\boldsymbol D_i$, $\boldsymbol M_i$ denote the $i$-th depth map and valid foreground (i.e., tissue) mask, respectively, $\boldsymbol K$ means the intrinsic matrix, $\odot$ means the element-wise multiplication. By default, the first frame is selected to initiate the Gaussian point cloud as the canonical state, i.e., ${\boldsymbol P_c}={\boldsymbol P}_{0}$. However, with the presence of tool occlusions in the colored image, some pixels in ${\boldsymbol I}_{0}$ are filtered out by $\boldsymbol{M_0}$, resulting in voids and local sparsity on the initialized point cloud.
% However, with the presence of tool occlusions and lighting reflections in the colored image, some pixels in ${\boldsymbol I}_{0}$ with invalid color supervisions or depth estimations are filtered out by $\boldsymbol{M_0}$, resulting in voids and local sparsity on the initialized point cloud. 
This unevenly distributed initialization consumes more time densifying the point cloud and leads to a sub-optimal efficiency. 
% Meanwhile, densely fusing 3D point clouds from all the frames risks impairing canonical geometry and drastically increasing the memory cost. Holding an assumption that dense Gaussian point distribution facilitates reconstruction on intensively deformed regions of the surgical scene,
Holding an assumption that dense Gaussian point distribution facilitates reconstruction on intensively deformed regions of the surgical scene, we develop a Motion-Aware Point Fusion (MAPF) scheme to selectively fuse points exhibiting intensive motions. 
% Specifically, given camera poses $\boldsymbol T$, we re-project all the frames into the canonical view (i.e., ${\boldsymbol I}_{i\rightarrow0}$) by:
% \begin{align}
% {\boldsymbol P}_{i\rightarrow0}\leftarrow\boldsymbol T_0^{-1}\;{\boldsymbol T}_i\;{\boldsymbol P}_i\;\boldsymbol,\;\;{\boldsymbol I}_{i\rightarrow0}\leftarrow\boldsymbol K{\boldsymbol P}_{i\rightarrow0}.
% \end{align}
Specifically, a motion-aware occlusion mask $\boldsymbol{F}$ is computed by combining occluded regions and pixels with large color differences from a pixel-wise-averaged image as following:
\begin{align}
    \boldsymbol{F} = \boldsymbol{\mathbb{I}}(\;\vert \boldsymbol{C}_{0}-{{\textstyle\sum_j^N}\boldsymbol{C}_{j}}/N\vert >\tau) \cup (\boldsymbol{1}-\boldsymbol{M}_0),
    \label{F}
\end{align}
where $\mathbb{I}(\cdot)$ refers to the indicator function, $\boldsymbol{C}_i$ represents the colored image of the $i$-frame, $N$ is the total number of frames, and $\tau$ is the threshold determining pixels with significant motions. $\boldsymbol{F}$ masks regions with large motions and local sparsity of the initialized Gaussian points ${\boldsymbol P}_{c}$ (i.e., ${\boldsymbol P}_{0}$). Finally, 3D points of $\boldsymbol{P}_{i}$ with 2D projected pixels in mask $\boldsymbol{F}$ will be fused with the $\boldsymbol{P}_c$ to initialize the canonical state as shown in Fig.~\ref{framework}.
\subsection{Optimization}
Our proposed framework jointly optimizes the canonical Gaussians $\mathcal {G}_c$ and the deformation model given by $\boldsymbol\Theta$. Given the tissue mask $\boldsymbol M$, we train our deformable tissue reconstruction framework by supervising the rendered images and depths by ground-truth colored images and stereo depth maps as following:
\begin{align}
    {\mathcal L}_C &=\vert\vert \boldsymbol M\odot(\boldsymbol {\widehat C}-\boldsymbol C)\vert\vert,\;\;{\mathcal L}_D=\vert\vert \boldsymbol M\odot(\boldsymbol {\widehat D}^{-1}-\boldsymbol D^{-1})\vert\vert_{},
\end{align}
where the $\boldsymbol {\widehat C}$, $\boldsymbol {\widehat D}$, $\boldsymbol { C}$, and $\boldsymbol { D}$ denote the rendered image, rendered depth, GT image, and stereo depth, respectively. The overall training loss is summarized  ${\mathcal L} = {\mathcal L}_C + {\mathcal L}_D$.

\section{Experiment}
\subsection{Experiment Setting}
\noindent\textbf{Datasets and Evaluation.}
We evaluate the proposed method and compare it with existing works on two datasets: 1) EndoNeRF dataset~\cite{wang2022neural} is a collection of stereo endoscopic videos including 6 clips extracted from Da Vinci robotic prostatectomy data. Each clip 
 is captured from a single camera viewpoint with complex surgical instrument occlusion and tissue deformations. 2) StereoMIS dataset~\cite{hayoz2023pose} is a stereo endoscopic video dataset captured from in-vivo porcine subjects containing diverse anatomical structures and challenging scenes with large tissue deformations. Specifically, we use all 6 scenes of EndoNeRF and select 3 clips from video P2\_7 and P3 in StereoMIS datasets with more diverse anatomical structures compared to EndoNeRF dataset. Each selected clip lasts for $4\sim5s$ with 30 $fps$. Following~\cite{zha2023endosurf}, we split frames of each scene into training and testing sets with a ratio of 7:1. We use PSNR, SSIM, and LPIPS to evaluate the reconstruction performance. Also, training time and rendering speed are calculated to evaluate the efficiency.
 
\noindent\textbf{Implementation Details.}
For each scene, we normalize the video duration into $[0,\;1]$ and empirically apply 17 learnable Gaussian basis functions to compose FDM. The training lasts for 3000 iterations, with an initial learning rate of $1.6\times10^{-3}$. To stabilize the training, we freeze the densification on Gaussian points number at the initial 600 iterations. All the experiments are based on the PyTorch framework~\cite{imambi2021pytorch} and conducted with a single NVIDIA RTX A5000 GPU.  

\subsection{Comparison with State-of-the-art Methods}
\begin{table}[t]
\centering
% \scriptsize
\resizebox{\textwidth}{!}{
\begin{tblr}{
    % width=0.4\textwidth, % Adjust the overall width of the table
    % colspec={X[0.8,l]X[1.3,c]X[0.8,c]X[1,c]X[0.8,c]X[1.3,l]X[1.3,l]}, % Adjust the width of each column here
    cells={c},
    cell{2}{1}={r=4}{},
    cell{6}{1}={r=4}{},
    % vline{2}={-}{0.08em},
    vline{3}={-}{0.08em},
    vline{6}={-}{0.08em},
    hline{1-2,6,10}={-}{0.08em},
    row{1-10} = {rowsep=0pt}, % Set the height of all rows except the header row to zero
    row{1} = {abovesep=0.1em}, % Set the height of the header row
    row{2} = {belowsep=0.1em}, % Set the height of the second row
}
Dataset   &  Method       & PSNR$\uparrow$  & SSIM(\%)$\uparrow$  & LPIPS$\downarrow$ & Time (sec)$\downarrow$                   & Speed ($fps$)$\uparrow$   \\
EndoNeRF  & EndoNeRF     & 35.55 & 93.02 & 0.09  & $\sim$21600    & 0.03   \\
          % & EndoSURF     & 36.47 & 95.23 & 0.08  & $\sim$600    & 0.03   \\
          & LerPlane     & 36.56 & 94.36 & 0.07  & $\sim$600 & 1.45   \\
          & EndoGaussian & \underline{37.66} & \textbf{95.89}& \underline{0.06}  & \underline{138}             & \underline{128.13} \\
          & Ours         & \textbf{37.90} & \underline{95.84} & \textbf{0.06}  & \textbf{64}          & {\textbf{338.80}} \\
StereoMIS & EndoNeRF     & 28.86 & 74.15 & 0.27  & $\sim$21600   & 0.03   \\
          % & EndoSURF     & 29.62 & 82.03 & 0.24  & $\sim$600    & 0.03   \\
          & LerPlane     & 29.46 & 77.73 & \textbf{0.20}  & $\sim$600 & 1.52   \\
          & EndoGaussian &  \underline{30.25} & \textbf{82.75} & 0.21  & \underline{151}              & {\underline{134.50}}  \\
          & Ours         & \textbf{30.48} & \underline{82.74} & \underline{0.21}  & \textbf{66}              & \textbf{330.37}\\
          
\end{tblr}
}
\label{SOTA}
\caption{Quantitative evaluation of our proposed framework against existing methods on endoscopic scene reconstruction. `Time' and `Speed' denote the training time and rendering speed ($fps$), respectively. The optimal and suboptimal results are shown in  \textbf{bold} and  \underline{underlined} respectively.}
\end{table}

% \begin{table}[t]
% \centering
% \arrayrulecolor{black}
% \resizebox{0.5\textwidth}{!}{
% \begin{tabular}{>{\centering}p{2.8cm} >{\centering}p{1.5cm} >{\centering}p{1.6cm} >{\centering}p{1.5cm} >{\centering}p{2cm} >{\centering\arraybackslash}p{2.2cm}} 
% \hline
% Method           & PSNR$\uparrow$  & SSIM(\%)$\uparrow$  & LPIPS$\downarrow$ & Time$(sec)\downarrow$  & Speed $(fps)\uparrow$    \\ 
% \hline
% Ours-HexPlane   & 37.13 & 95.51 & 0.06  & 108     & 182.5  \\
% Ours-PFS         & 37.08 & 95.29 & 0.07  & 60      & 338.1  \\ 
% \hline
% w/o MAPF   & 37.47 & 95.62 & 0.07  & 61     & 348.8  \\ 
% Ours             & 37.90 & 95.84 & 0.06  & 64  & 338.8  \\
% \hline
% \end{tabular}}
% \caption{Quantitative analysis of the key components on EndoNeRF dataset. `Time' and `Speed' denote the training time and rendering speed, respectively.}
% \arrayrulecolor{black}
% \end{table}
% \begin{table}[]
% \begin{tabular}{l|ccc}
% \hline
% Method       & PSNR$\uparrow$  & SSIM(\%)$\uparrow$  & Time(min)$\downarrow$ \\ \hline
% LerPlane     & 22.75 & 82.13 & ~1      \\
% EndoGaussian & 35.81 & 94.67 & ~1      \\
% Ours         & 37.90 & 95.84 & ~1      \\ \hline
% \end{tabular}
% \end{table}

\begin{table}[h]
    \begin{minipage}{0.45\linewidth}
        \centering
        \resizebox{\textwidth}{!}{
        \begin{tabular}{l|ccc}
        \hline
        Method       & PSNR$\uparrow$  & SSIM(\%)$\uparrow$  & Time($min$)$\downarrow$ \\ \hline
        LerPlane     & 22.75 & 82.13 & $\sim$1      \\
        EndoGaussian & 35.81 & 94.67 & $\sim$1      \\
        Ours         & \textbf{37.90} & \textbf{95.84} & $\sim$1      \\ \hline
\end{tabular}}
\label{comparison_1min}
\caption{Quantitative comparison with SOTAs given the limited training time around 1 $min$.}
    \end{minipage}
    \hspace{0.01\linewidth}
    \begin{minipage}{0.55\linewidth}
        \centering
        \resizebox{\textwidth}{!}{\begin{tabular}{>{\centering}p{2.8cm} >{\centering}p{1.5cm} >{\centering}p{1.6cm}  >{\centering\arraybackslash}p{2cm} }
        % >{\centering\arraybackslash}p{2.2cm}} 
\hline
Method           & PSNR$\uparrow$  & SSIM(\%)$\uparrow$   & Time$(sec)\downarrow$      \\ 
\hline
Ours-HexPlane   & 37.13 & 95.51& 108    \\
Ours-PFS         & 37.08 & 95.29   & \textbf{60}      \\ 
\hline
w/o MAPF   & 37.47 & 95.62  & 61       \\ 
Ours             & \textbf{37.90} & \textbf{95.84}   & 64  \\
\hline
\end{tabular}}
\label{ablation}
\caption{Quantitative analysis of the key components on EndoNeRF dataset. `Time' denotes the training time.}
    \end{minipage}
\end{table}
% \begin{table}
% \centering
% \begin{adjustbox}{height=3cm, center}
% % \resizebox{\columnwidth}{!}{
% \begin{tblr}{
% width=0.8\textwidth,
% % hline{2-Y} = 1pt, vline{2-Y} = 1pt,
%   cells = {c},
%   cell{2}{1} = {r=5}{},
%   cell{7}{1} = {r=5}{},
%   vline{6} = {-}{0.08em},
%   hline{1-2,7,12} = {-}{0.08em},
% }
% Dataset   & Method       & PSNR $\uparrow$  & SSIM $\uparrow$  & LPIPS $\downarrow$ & Time $\downarrow$                   & FPS $\uparrow$   \\
% EndoNeRF  & EndoNeRF     & 35.55$\pm$1.3 & 93.02 & 0.09  & $\sim$ 6 $h$     & 0.03   \\
%           & EndoSURF     & 36.47 & 95.23 & 0.08  & $\sim$ 10 $h$    & 0.03   \\
%           & LerPlane     & 36.56 & 94.36 & 0.07  & $\sim$ 10 $min$ & 1.45   \\
%           & EndoGaussian & 37.66 & 95.89 & 0.06  & 2.3 $min$                 & 128.13 \\
%           & Ours         & 37.90 & 95.84 & 0.06  & 1.1 $min$                 & 334.5  \\
% StereoMIS & EndoNeRF     & 28.86 & 74.15 & 0.27  & $\sim$ 6 $h$     & 0.03   \\
%           & EndoSURF     & 29.62 & 82.03 & 0.24  & $\sim$10 $h$    & 0.03   \\
%           & LerPlane     & 29.46 & 77.73 & 0.20  & $\sim$10 $min$ & 1.52   \\
%           & EndoGaussian & 30.25 & 82.75 & 0.21  & 2.5 $min$                 & 134.5  \\
%           & Ours         & 30.48 & 82.74 & 0.21  & 1.1 $min$                 & 330.37 
% \end{tblr}
% \end{adjustbox}
% \end{table}

\begin{figure}[t]
    \centering
    \includegraphics[width=0.95\textwidth]{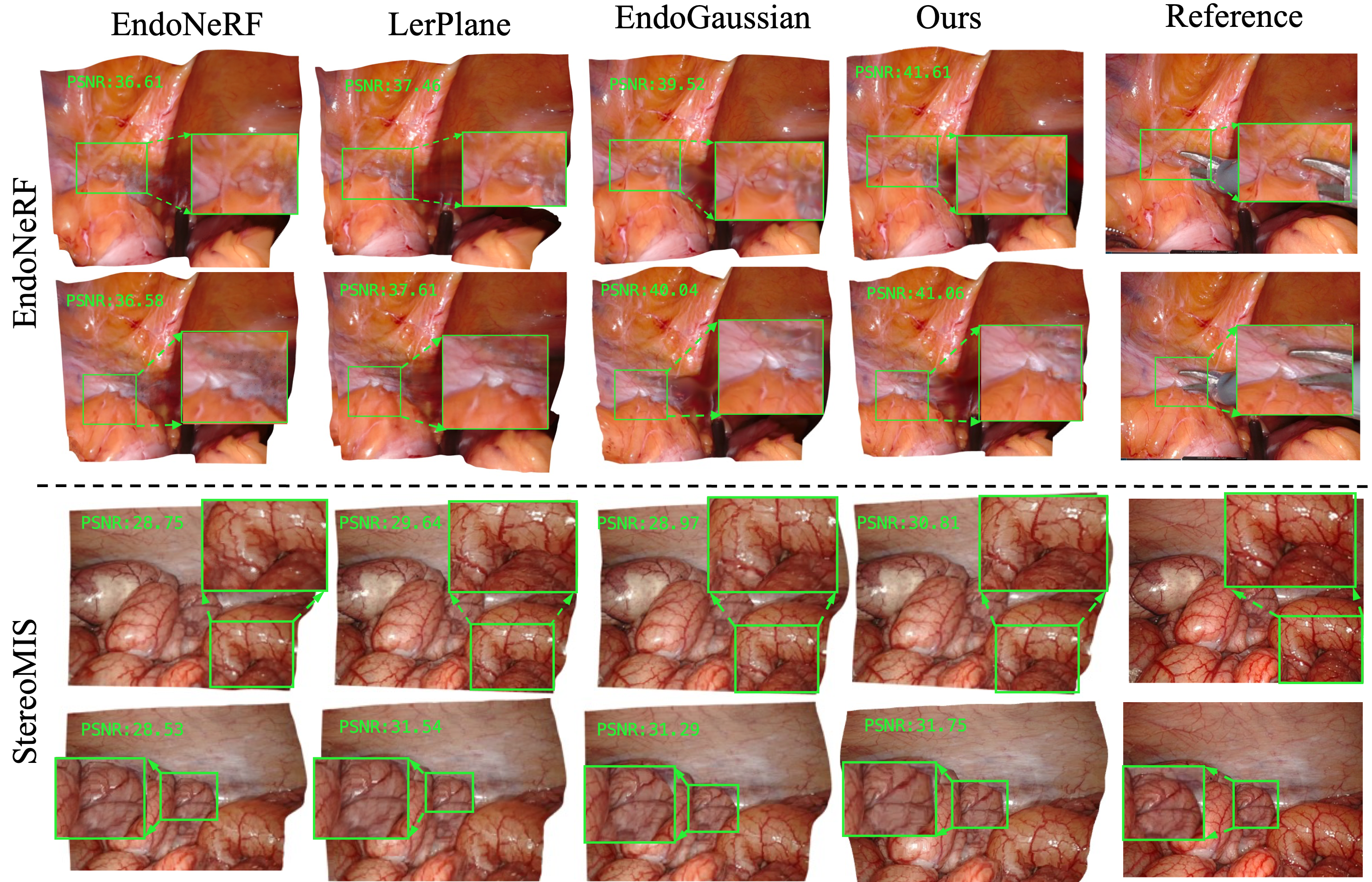}
    \caption{Visualization of the 3D reconstruction results. }
    \label{visualization}
\end{figure}
We evaluate our proposed framework by comparing its performance with EndoNeRF~\cite{wang2022neural} and 
 two existing SOTAs on fast reconstruction: LerPlane~\cite{yang2023neural}, and EndoGaussian\cite{liu2024endoG}. EndoGaussian is a concurrent work using GS to accelerate the endoscopic reconstruction. As listed in Table~1, despite the effectiveness of reconstructing deformable tissues, EndoNeRF takes a long training time (hours) to reconstruct a scene in seconds, which compromises their intraoperative usability. As a comparison, the fast reconstruction method, LerPlane, effectively accelerates the training phase to the minute level and leads to a superior reconstruction quality. Our framework yields noticeable performance gains over LerPlane across all evaluation metrics on reconstruction quality, 
%$\uparrow$ PSNR, $\uparrow$ SSIM, and $\downarrow$ LPIPS, 
while substantially improving training efficiency by 10 times to around 1 minute. Furthermore, we compare our method against EndoGaussian~\cite{liu2024endoG}. Similar to~\cite{wu20234d}, this method relies on decomposed feature planes to model dynamic Gaussian Splatting, presenting remarkable performance against NeRF-based methods and reaching comparable reconstruction quality to our framework. However, benefitting from the highly efficient FDM, our method leads to over 2$\times$ acceleration in both training (138 $sec$ $\rightarrow$ 64 $sec$) and rendering (128 $fps$ $\rightarrow$ 338 $fps$). Note that our measured training and rendering speeds for EndoGaussian mismatch with the reported values in~\cite{liu2024endoG} due to different hardware used.

Despite the comparable reconstruction quality, as shown in Table~2, our proposed method achieves significantly superior performance within a limited training time, which demonstrates the superiority in intraoperative scenarios.

We also visualize several rendered scenes as shown in Fig~.\ref{visualization} for better qualitative evaluation. It can be observed that our method has an enhanced capability of preserving the appearance details and modeling complex tissue motions. Additionally, rendered views given by EndoGaussian indicate comparable rendering quality to ours without a visually perceivable difference. According to these results, our proposed method achieves state-of-the-art (SOTA) performance on endoscopic scene reconstruction, especially its outstanding progress in fast training and real-time-level rendering, which indicates great clinical values in intraoperative applications.

\subsection{Quantitative Evaluation of Key Components}
We first investigate the effectiveness of the proposed FDM by comparing it with existing deformation modeling techniques on the EndoNeRF dataset. With the identical workflow shown in Fig.~\ref{framework}, we replace the FDM with other modeling methods including a combination of Fourier and Polynomial series (Ours-FPS), and a HexPlane-based decomposed feature plane~\cite{cao2023hexplane} following~\cite{wu20234d} denoted as Ours-HexPlane. As illustrated in Table~3, despite the acceleration, Ours-FPS shows limited capability of representing complex deformations. On the other hand, using HexPlane to encode spatial and temporal information significantly enhances the representative capability, however, leading to a relatively longer training time. Our proposed FDM achieves the best deformation representing performance without impairing time efficiency.
% To better understand the properties of deformation fields, we visualize the trajectories of some example Gaussian points driven by different deformation models. As shown in Fig.~\ref{}, both Ours-HexPlane and the proposed GDDM capture the detailed deformations while Ours-FPS and Ours-PS tend to model over-smooth trajectories. 
Furthermore, `w/o MAPF' refers to initializing the Gaussian points with only the first frame point cloud, which exhibits a performance drop and thus demonstrates the important role of the proposed MAPF scheme.

\section{Conclusion}
In this paper, we work toward intraoperative surgical scene reconstruction by proposing a fast and accurate deformable scene reconstruction framework. With the utilization of Gaussian Splatting, our framework can achieve high-quality rendering at a real-time level. To further accelerate the tissue motion modeling, we introduce an efficient flexible deformation modeling scheme composed of learnable Gaussian basis functions to maintain a strong motion-representative capability. Besides, combined with a motion-aware point fusion scheme for initialization, our framework leads to a SOTA reconstruction quality while significantly minimizing the training time to only 1 $min$/scene, showing the possibility of reconstructing longer sequences with more challenging surgical scenes. Therefore, we believe that our work achieves significant progress in bridging the gap between high-quality rendering and intraoperative applications.

\section*{Acknowledgement}

This work was supported by Ministry of Education Tier 1 Start up grant, NUS, Singapore (A-8001267-01-00); Ministry of Education Tier 1 grant, NUS, Singapore (A-8001946-00-00).

\bibliographystyle{SPLNCS04} % We choose the "plain" reference style
\bibliography{main.bib} 
\end{document}